# Rule Generalisation using Snort

## U Aickelin, J Twycross and T Hesketh-Roberts

School of Computer Science and IT

University of Nottingham

NG8 1BB UK

E-mail: uxa@cs.nott.ac.uk, jpt@cs.nott.ac.uk, tmhesket@fish.co.uk

**Abstract:** Intrusion Detection Systems (IDSs) provide an important layer of security for computer systems and networks. An IDS's responsibility is to detect suspicious or unacceptable system and network activity and to alert a systems administrator to this activity. The majority of IDSs use a set of signatures that define what suspicious traffic is, and SNORT is one popular and actively developing open-source IDS that uses such a set of signatures known as SNORT rules. Our aim is to identify a way in which SNORT could be developed further by generalising rules to identify novel attacks. In particular, we attempted to relax and vary the conditions and parameters of current SNORT rules, using a similar approach to classic rule learning operators such as generalisation and specialisation. We demonstrate the effectiveness of our approach through experiments with standard datasets and show that we are able to detect previously undetected variants of various attacks.

## 1 Introduction

Computer attacks, e.g. the use of specialised methods to circumvent the security policy of an organisation, are becoming more and more common. IDSs are installed to identify such attacks and to react by usually generating an alert or blocking suspicious activity.

IDSs come in many forms which we overview in the following section. The work presented here is based on a popular network intrusion detection system (NIDS) called SNORT (2006). SNORT detects attacks by comparing live Internet traffic against signatures that define known attacks. SNORT is an open-source GNU (2006) NIDS and an example of a system that uses signatures, in this case known as SNORT rules. The aim of this paper is to determine the effectiveness of generalisation when applied to the matching of Internet traffic against SNORT's rule signatures.

In this paper we introduce a novel rule generalisation operator for creating new rules. In particular, we present two generalisation operators, invert and content, which can be used to either generalise or specialise SNORT rules. Analysing the results found by these new rules generates an improved understanding of attack patterns. Subsequently, better rules can be created beyond the classic learning operators based on blind addition, deletion or negation of rule conditions as first suggested by Mitchell (1997).

In the next section, we will talk about the current state of the art in IDS and highlight some potential shortcomings. We will then go on to explain SNORT and SNORT rule generalisation in section 3 and 4. Details of our system are presented in section 5 and results using real-world data are in section 6. Finally, the paper concludes by discussing the effectiveness and appropriate use of our rule generalisation in IDS signature processing.

## 2 Current State-of-the-Art in IDS

According to Crothers (2003), intrusion detection technology is technology designed to monitor computer activities for the purpose of finding security violations. An IDS is a system that implements such technology. The meaning of a security violation will vary between set-ups. To some, the definition of a security violation may be limited to activities breaching confidentiality and/or resulting in downtime of services. Generally, a security violation would be any deliberate activity that is not wanted by the victim. This typically includes denial of service attacks, port scans, gaining of system administrator access and exploiting system security holes such as the processing HTML forms on the server.

IDSs come in many different forms, and their method of finding security violations varies. Following Northcutt (2002), one division is often made in terms of IDS placement: Host-based (HIDSs) that detect attacks by analysing system logs or Network-based (NIDSs) that detect attacks by directly analysing network packets in real-time, e.g. Snort.

Here we will concentrate on misuse detection NIDSs. Techniques used by such NIDSs still have a lot of room to evolve. Northcutt (2002), Ning and Xu (2004) and Kim et al (2007) identify a number of problems associated with current misuse NIDSs:

- They cannot fully detect novel attacks;

- Variations of known attacks are not fully detected;

- They generate a large amount of alerts, as well as a large number of false alerts;

- Existing IDSs focus on low-level attacks or anomalies and do not identify logical steps or strategies behind these attacks.

Our work here mainly focuses on the second point. Signature sets are not effective against varied attacks if they are written to identify precisely each currently known issue. Conversely, using signatures with more general matching criteria results in a higher proportion of legitimate network traffic generating false alerts. We address this issue by systematically implementing generalised rules and alerts.

There are a number of alternative methods of identifying new or variations of known attacks that are currently under investigation. The interested reader is referred to Axelsson (2000), who offers a survey of these techniques. Here we briefly summarise two related areas:

Gomez et al (2003) and Esponda et al (2004) use ideas based on the Human Immune System to build an artificial immune system. The artificial immune system is then used to evolve competitively new rule sets. This allows the generation of rules that characterize the non-self space (abnormal) by just taking self (normal) samples as input. The difference to our work is that we use crisp or fixed rules derived by generalising SNORT rules.

The scenario approach Ning and Xu (2004) addresses the problems of large amounts of alerts and lack of attack strategy consideration by proposing correlation of related alerts. The principle is that certain attacks would have a likely prerequisite, such as scanning for the existence of an open port before attacking it. In this way, for example, the port scan and the attack may be be correlated into a single alert as part of the same attack process. Burgess (2006) also uses statistical methods. In his case a filter based on a time-series prediction detects the significance of deviation. The extent of the deviation determines how the system should respond. However, research into these correlations still in the beginning as corrrelating attacks is often neither obvious nor easy.

## 3 Snort and Snort Rules

SNORT is one of the most popular NIDS. SNORT is Open Source, which means that the original program source code is available to anyone at no charge, and this has allowed many people to contribute to and analyse the programs construction. SNORT uses the most common open-source licence known as the GNU General Public License. Recent research issues addressed with SNORT include alert visualisation by Hoagland and Staniford (2003) and automated port-scan detection by Staniford et al (2002).

Lawton (2002) discusses the advantages and disadvantages of security software being open-source. In the article, Lawton introduces the argument that the availability of open source software code makes it easy for hackers to figure out how to defeat the security. Lawton also weighs up the counter-argument that closed-source security systems are still compromised and that code being open-source allows security holes to be closed as soon as they are identified, as well as enabling code to be customised for individual security needs. On balance, we believe that open-source is an advantage for computer security.

SNORT, like most NIDSs, uses a set of signatures to define what constitutes an attack. SNORT signatures are regularly updated on the SNORT website, usually several times a day, which can be confirmed by periodically checking the timestamps next to available downloads SNORT. SNORT is flexible in how it can be utilised, as (Figure 1) begins to demonstrate. A file containing previously logged traffic can be used as input to SNORT, in exactly the same way as live traffic. SNORT also supports a range of outputs, such as saving alerts to files or databases, or creating a network traffic log of all received traffic for later processing in the case of live traffic capture. The flexibility exists for SNORT to support virtually any output method, due to an ability to support both in-house and third party output plug-ins.

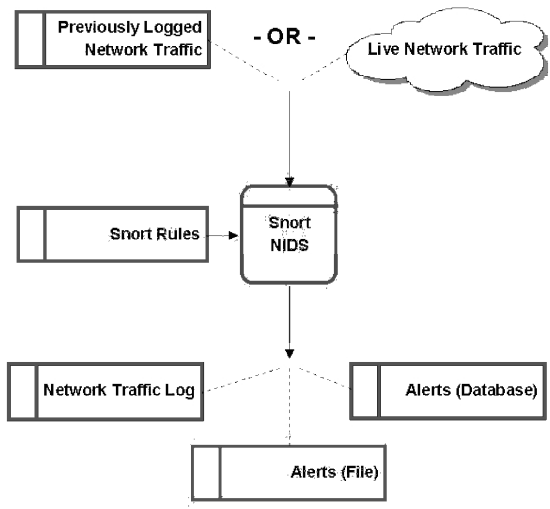

Figure 1: Data-Flow Diagram demonstrating the flexibility in utilising SNORT.

Since a clear understanding of SNORT's rules is crucial for our research, a detailed explanation follows. A summary of this information can be found in Table 1 below. SNORT's signature sets, which are used for identifying security violations, are called SNORT rules. Groups of SNORT rules are referred to as a .rules file, each of which can be selectively included into the SNORT configuration file snort.conf. A .rules file is a plain text file in which each line holds a separate rule. The following notes on SNORT's rule format were put together using the SNORT Users Manual, for full detail see SNORT's website.

A rule is formally defined as shown below. Text in <angled brackets> would be replaced with the appropriate compulsory variable, without angled brackets present. Text in [square brackets] is optional and either represents nothing or represents the text itself, in either case without the square brackets.

```
<rule action> <protocol> [!]<source ip> [!]<source port>
<direction> <dest ip> <dest port> <rule options>

example: alert tcp any any -> any 25
[create an alert for any incoming traffic send to port 25]
```

The term rule action describes what response is made in cases when the conditions in the rule match when compared against an Internet packet. Most commonly, the rule action is alert, which usually means saving alert data to a file or database for later retrieval or for another application to process. Alert generating packets are also logged. Another action includes log, useful when it is inappropriate to generate an alert, but the traffic is of some interest. Other actions include pass (allow the packet through) and activate (start other rules or actions).

For the protocol and port statements, please refer to SNORT(2006). For the IP statements, SNORT uses an IP/CIDR (Classless Inter-Domain Routing) block number after the IP address (see Fuller et al (1993)). The packet data must identify as coming from or going to the IP address range given. An optional exclamation mark can be placed in front of the IP address to invert the meaning of the rule. Values can be given as ranges of IP/CIDR statements, e.g. 192.168.1.0/24. All possible IP addresses can be represented by using the keyword any.

The packet data must identify itself as coming from (going to) the Internet port (or port range) given (the Source/Destination Port statements). An optional exclamation mark can be placed in front of the port statement to invert the meaning of the rule. A specific port number or a range can be stated by using a colon (:) to separate the lowest and the highest port number, e.g. 1:1024. Alternatively, all possible port numbers can be represented by using the keyword any. The direction statement specifies whether the packet is from the source to the destination or vice versa.

There are also additional rule options, including further conditions for the rule to match, the message to be used in alerts and options for activate rules. The interested reader is referred to the SNORT website for more details. Rule options are separated from each other using semi-colons. Some options have a parameter value associated with them, in which case a colon separates the option name and option value.

## 4  Rule generalisation

We propose to generate new rules by generalising SNORT rules. Given an Internet packet that contains a variation of a known attack, there should be some automated way to identify the packet as nearly matching a NIDS attack signature. If a particular statement has a set of conditions against it, an item may match some of the conditions. Whereas Boolean logic would give the value false to the query 'does this item match the conditions', our logic could allow the item to match to a lesser extent rather than not at all. This principle can be applied when comparing an Internet packet against a set of conditions in a SNORT rule. Our hypothesis is that if all but one of the conditions are met, an alert with a lower priority can be issued against the Internet packet, as the packet may contain a variation of a known attack.

In our implementation, generalisation in the case of matching network packets against rules, involves allowing a packet to generate an alert if:

- The conditions in the rule do not all match, yet most of them do;

- The only conditions that do not match exactly nearly match.

As an example, assume a certain rule states that an alert should be generated if a packet is a particular length, on a particular port and contained a certain bit pattern. Using our generalisation a packet matching those criteria, except perhaps on a different port, or with a slightly different bit pattern, would still count as matching, and a (modified) alert would be generated.

# 5  Implementation

Our implementation is made up of three components (Figure 2):

- The first program, called FuzzRule, processes the Snort rules (.rules files) and creates two new sets of rules using two generalisation principles (Invert and Content);

- The second program, AlertMerge, merges alert files generated from the original rules with alert files generated from the generalised rules;

- The third program, the FuzzRule post-processor, summarises alerts given. By checking this summary, we can identify where large numbers of false positives are being generated. Thus, we can adjust FuzzRule to reduce false alert rates.

## 5.1  FuzzRule

The FuzzRule program, which Figure 2 provides a diagrammatic overview of, meets the following specifications:

Given a .rules file, the application saves a back-up of the original file before replacing it. For each SNORT rule in the original .rules file, the application includes the original rule in the new .rules file and follows this with each variation of the rule generated using our generalisation. Each generalised variation of an original rule is generated either by inverting or removing the meaning of one of the rule parameters. Thus, when comparing the property of a packet against the generalised rule, the packet should match all cases that are similar to the original rule. As we will see in the next section, there is a difference between removing and inverting, and the correct behaviour can only be achieved by inverting rule options.

Based on an initial set of experiments, we identified the following rule parameters as being good candidates for generalisation. Any of these present in a rule will be generalised using the stated method (more details later in this section):

Inversion: Port, IP address, Direction, Protocol, Content, URI Content;

- Special Inversion: Depth, Offset;

- Generalisation of Content: Content, URI Content;

- Both original rules and generalised variations of rules have their alert message tagged so that it can be identified in what way a matching rule has been generalised (if at all);

- A program option is provided giving each generalised variation of a rule a lower priority setting. A priority is a numerical value from one upwards, one being the highest priority and representing the most severe attack and any larger number being less severe. The priority setting is not used by Snort, but serves as an indicator for the operator browsing the alert file.

*Title*

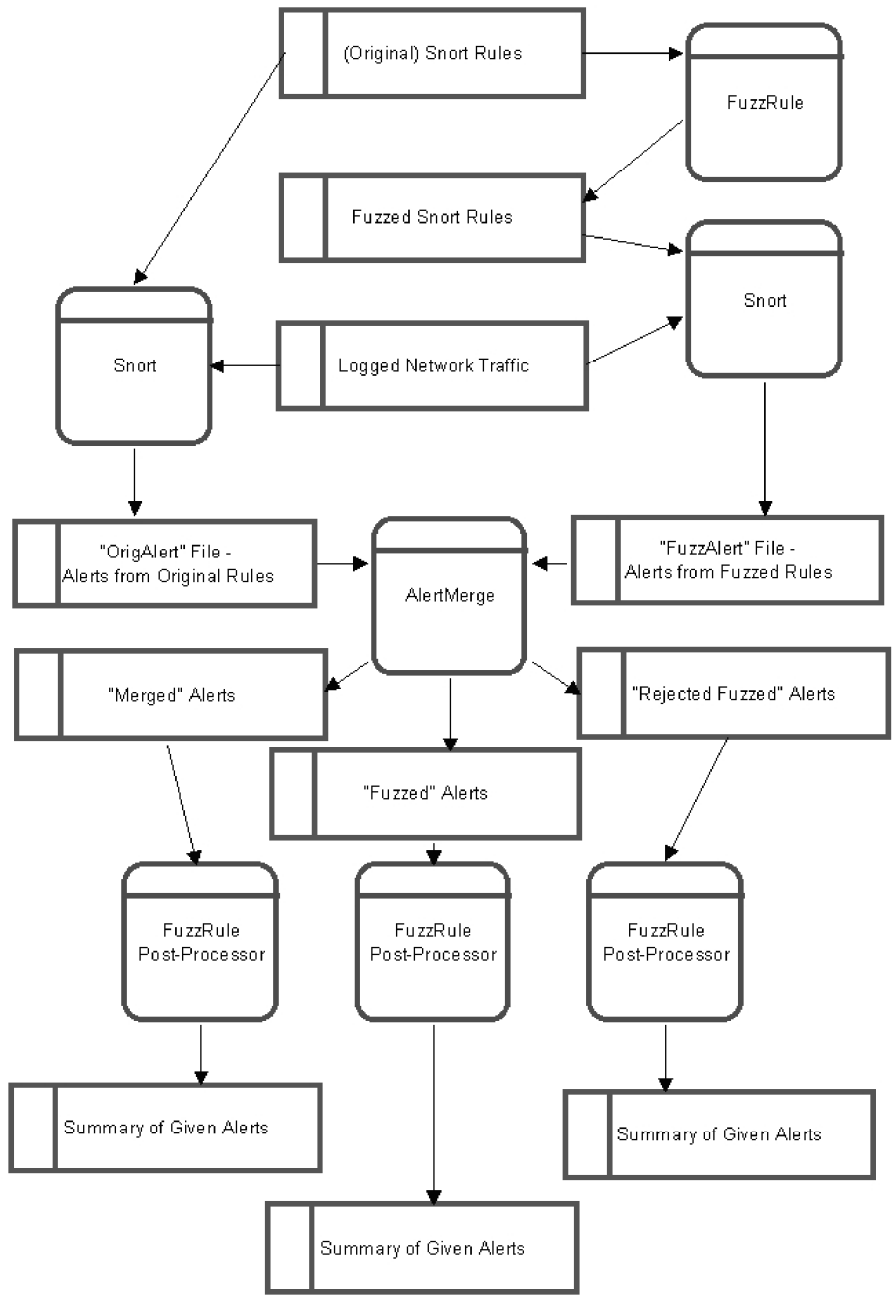

Figure 2: Data-Flow Diagram for Overall System.

*5.2 Generalisation by Rule Inversion*

When designing our program, generalisation was at first applied by removing a single rule option to form a generalised rule. For example, given the original rule below, one generalised variation is given afterwards. By removing the offset parameter, more packets will match against the generalised rule than the original rule:

```
alert udp any any -> any 69 (msg:TFTP GET Admin.dll;
content:
|0001|; offset:0; depth:2; content:admin.dll; offset:2;
nocase; classtype:successful-admin; reference:url,
www.cert.org/advisories/CA-2001-26.html; sid:1289; rev:2;)

alert udp any any -> any 69 (msg:TFTP GET Admin.dll;
content:
|0001|; offset:0; content:admin.dll; offset:2;
nocase; classtype:successful-admin; reference:url,
www.cert.org/advisories/CA-2001-26.html; sid:1289; rev:2;)
```

Using the above removal generalisation principle means that if a packet matches an original rule, it typically also matches all generalised variations of the same rule. By design, SNORT produces at most one alert per packet. When we first tried the removal approach, we expected that SNORT matched the original rule by default, due to it appearing before the generalised variations in the .rules file. However, during run-time tests, alerts were only generated from generalised rules. Closer investigation revealed that SNORT places rules into an efficient binary tree-style system for quicker processing and traverses the tree by matching lower-cost matching rules first. Thus, rules with fewer options, like our removal generalised rules matched before their original counterparts. Changing the priority and/or SNORT id of the alerts cannot change this behaviour in any way.

Therefore, a new principle had to be applied. Instead of removing rule options, we inverted them. The principle of inverting a rule option is defined as matching a packet in only those cases that are similar but where the original rule option would not have matched.

The difference between the removal generalisation principle and the inversion generalisation principle is made clear in Table 1 using a rule with four conditions A, B, C and D. The removal principle means that only if not all original rule conditions hold, a maximum of one generalised rule matches. The same is true of the inversion principle. However, if all original rule conditions match, all generalised rules under the removal principle will also match. In contrast, under the inversion principle, if all the original rule conditions match, none of the generalised rules will. The latter is the desired outcome and hence our choice for implementation.

| Rule | Removing | | | | Inverting | | | |
|------|---|---|---|---|------|------|------|------|
| Original | A | B | C | D | A | B | C | D |
| generalised 1 | A | B | C | - | A | B | C | not D |
| generalised 2 | A | B | - | D | A | B | not C | D |
| generalised 3 | A | - | C | D | A | not B | C | D |
| generalised 4 | - | B | C | D | not A | B | C | D |

Table 1: Different Generalisation Principles Demonstrated Using Conditions A-D.

Using the inversion principle, applying generalisation is straightforward for most rules, e.g. inverting ports, IP addresses, protocols, traffic direction or negating complete content or URI content strings. Unfortunately, for some rule options finding the generalised counterpart is more complicated. As an example, let us have a look how we created generalised versions of the depth and offset rule options.

The depth and offset options affect which part of the packet data the content option is matched. An offset value means that the content string is not compared against until an 'offset' number of bytes into the packet data. A depth value dictates how many bytes from the start of the offset (or start of the packet if no offset is given) a comparison between the packet data and the content string should be made for. The principles by which the depth and offset options are generalised are as follows:

- In the generalised variations of the rule, the region(s) of the packet header not compared against in the original rule, are compared against, meaning that it should find a match in some cases when it would not have done with the original rule.

- To compare packet data before the region currently being compared, all bytes should be compared prior to the offset, plus the length (minus 1) of the content string to match bytes into the offset. This maximises the chance to match what would not have previously been found because the content string could partially exist within and outside the region.

- To compare packet data after the region originally being compared, all bytes should be compared after depth characters from the start of the original search region, plus the length (minus 1) of the content string to match bytes into the end of the original search region. The same principle applies as in the offset case.

Finally, we need to discuss an effective method of content generalisation, since this is often the key to matching a rule against traffic patterns. The content option specifies a string to search for in packets. Applying generalisation to the content option means individual characters in the content are replaced with a question mark (?) to represent any character during a match. Additionally, the content option value can be shortened slightly, which could allow a match if start or terminating characters in the attack sequence differed. This type of generalisation is applied to all rules with a content (actual content) and uricontent (e.g. web addresses) option. In all cases a number of generalised rules are made by substituting one character in turn with a ?.

## 5.3 AlertMerge

The AlertMerge program is shown diagrammatically in Figure 3. The program accepts two alert files, both generated by SNORT against the same traffic. One of the alert files is generated using the original and the other generalised rules. These files are referred to as original alert and generalised alert files respectively, from now on. Three output files are generated, each one with the same file name as the alert file, but with a particular extension appended to the end of the name. One file with a .merged extension, containing some alerts from the generalised alert file and all alerts from the original alert file.

As discussed previously, SNORT may alert against a less vital generalised rule instead of an original rule if a packet matches both. Thus, the alert file from generalised rules alone may imply that the traffic is less severe than it really is. The merging process ensures that in a merged alert file only one alert per data packet is recorded. If two alerts, one from each of the two given alert files, are generated from the same packet, then only the alert from the original alert file is saved to the .merged file. The alerts are kept in chronological order. One file with a .fuzz extension, which contains all alerts generated by the generalised rules that were accepted into the .merged file. Finally, one file with a .rejected_fuzz extension, which contains all alerts from the generalised alert file that did not make it into the .merged file. This file is useful for identifying which generalised rules are being matched with precedence over original rules.

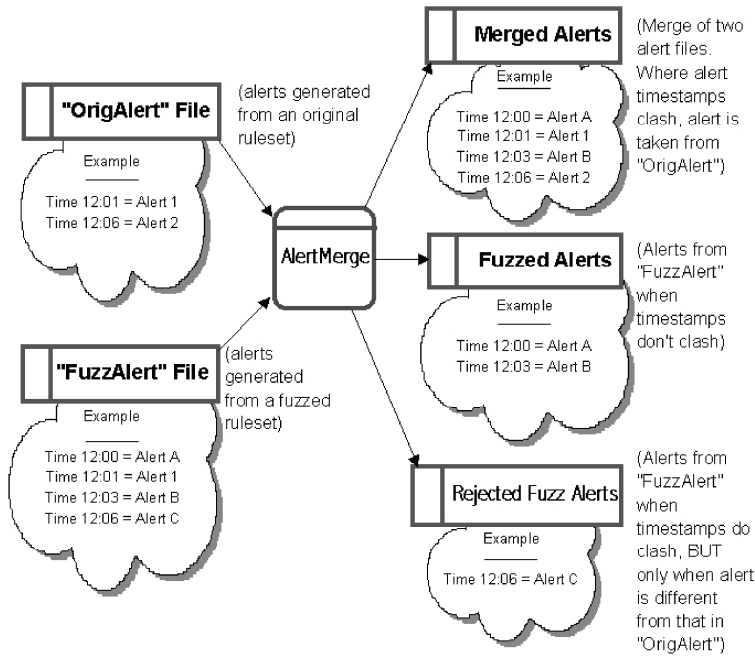

Figure 3: Data-Flow Diagram Overview of AlertMerge.

## 5.4 *FuzzRule Post-Processor*

The FuzzRule post-processor program, summarises the alert file generated by Snort. The file is summarised regardless whether it was taken directly from Snort, or whether the alert file was one of the four possible outputs from AlertMerge. Then an alert summary is given as the total number of occurrences of each alert and the total number of occurrences of each generalisation method implemented, including the number of occurrences of original alerts.

## 6 Experiments and Analysis of Results

This section reports upon program performance, as well as analysing how effective various attempts at applying generalisation have been. To ensure fair comparisons, all analysis will be performed under the same testing environment, which is a 2000 MHz AMD-powered PC with 1024Mb RAM running the Linux Mandrake 9.0 operating system and Snort 2.6.0.

When testing SNORT rules and alert files generated from them, tcpdump traffic data is used from Lincoln Laboratory IDS test data sets available from the Massachusetts Institute of Technology (MIT data 1999). Although we are aware of some of the limitations of these data sets mainly due to their age, they were chosen due to the deliberate mix of typical legitimate traffic (including consideration for different ports and client platforms), with attacks, both of a known and novel nature and of various levels of severity. Most importantly, these data sets are still the most realistic publicly available with a full list of actual attacks.

Tcpdump is a utility (generally available on most UNIX systems) that can save raw packet data to a tcpdump binary file. NIDSs such as SNORT can optionally process archived traffic data from tcpdump files rather than live traffic data. In this case, SNORT deals with the data in exactly the same way as if it were live (with the exception that tcpdump data is processed at the rate that the CPU allows, whereas live data must be processed real-time and is vulnerable to packets not being analysed by the NIDS if it can not keep up with processing).

The tcpdump file used for analysis in each case, was the outside.tcpdump file from Thursday, week 4 of the 1999 data sets. The 1999 data sets were chosen over 1998, since they reflected a more up-to-date range of attacks, and week 4 was chosen since this data contained a range of attacks among normal traffic for the very purpose of testing. The attacks contained in the test data are also listed by Lincoln Laboratory. We used this list to confirm that our systems found all previously known attacks in the data file.

When implementing generalised Inversion, the execution time was 1 second to process and convert the original 1,325 rules into a total of 6,975 rules. The generalised Content execution time was 2 seconds to process and convert the same 1,325 original rules, into a total of 18,265 rules. These execution times would easily be acceptable for most potential uses, such as each time the SNORT rules were downloaded for signature updates. The increase in the number of rules affected the time spent processing network traffic data as follows:

- Using the original rules, Snort took approx 100 seconds to process 1,635,267 packets;

- Using the generalised (inverted) rules, Snort took approx 400 seconds to process the same packets;

- Using the generalised content rules, Snort took approx 1,000 seconds to process the packets.

The change in SNORT's processing time is an increase of around four to ten times and roughly in line with the increase in the number of rules. We believe that such a processing time increase is not a problem and still well within real-time processing requirements. On our moderate system, a whole days worth of data is processed in less than two hours.

*6.1 Content rule generalisation*

In a second experiment, we used generalised content on the content and uricontent options. The generalised content principle produced far less generalised alerts. Out of 1,635,267 packets, only 50,081 packets or 3% generated an alert. Nearly 38,000 (more than 75%) of these alerts were generated against just one rule (WEB-MISC ICQ Webfront HTTP DOS). Thus, this rule can probably be ignored as creating false alarms.

A detailed further analysis of these results is more complex and beyond the scope of this paper. However, briefly one can note that generalising the content option (Types: cor,rx+) is responsible for a larger proportion of (probably false) alerts compared with the uricontent option (Types: urr,rx+). Ignoring the rule mentioned above, of the top four generalised alerts, three are generated from rules generalised by the content option, compared to just one by the uricontent option. As for inversion, in our opinion the most interesting cases are those appearing the least often, e.g. less than 10 occurrences.

Out of these, the most interesting are those appearing only four or six times. This makes it unlikely that these are false alerts. For instance, these unusual alerts matched when allowing for the destination port or destination IP address to be different from the original alert. This could be an indication that the particular SNORT rules that generated the alerts were too stringent in their criteria.

To reduce the number of false positives, we ignore those > 25, as they are very likely to be false or trivial alerts or already covered by the original SNORT rules.

| Rule ID | Frequency | Class | Found by original rules |
|---------|-----------|-------|-------------------------|
| 250 | 3 | False Alert | No |
| 255 | 3 | True Alert | No |
| 323 | 1 | Additional Information | Yes |
| 530 | 1 | Additional Information | Yes |
| 1201 | 10 | Additional Information | Yes |
| 1377 | 9 | FTP Beta Software Used | No |
| 1378 | 1 | Additional Information | Yes |

Table 2: List of Attacks Generated by Content Rule Generalisation

Let us have a look at some of the above in more detail to get a feeling for the usefulness of the generalisation. Generalised rule 250 gives a false alert: the content generalised gives false alerts as it places a wildcard against the only content character there is (see original rule below). Hence, we pick up any traffic to said port, which in almost all cases is harmless. A simple solution to this problem is to alter the rule generalisation algorithm to not allow content rule generalisation if there is only one content character.

```
Signature alert tcp $HOME_NET 15104 -> $EXTERNAL_NET any
(msg:"DDOS mstream handler to client";
flow:from_server,established; content:">";
reference:cve,2000-0138; classtype:attempted-dos; sid:250;
rev:4;)
```

Rule 255 finds new True Positives! These three packets were not picked up with the original rules. The generalised rule below alerts on the following three packets and associates them with the DNS zone transfer TCP attack. They are identified as attacks in the Lincoln Lab 'solutions' MIT data.

```
03/31wed-18:00:32.637334 194.7.248.153:2076 ->
172.16.112.20:53
04/02fri-15:53:24.050418 194.7.248.153:1238 ->
172.16.112.20:53
04/02fri-18:49:30.235173 195.73.151.50:7332 ->
172.16.112.20:53
```

Original Rule 255:

```
Signature alert tcp $EXTERNAL_NET any -> $HOME_NET 53
(msg:"DNS zone transfer TCP"; flow:to_server,established;
content:"|00 00 FC|"; offset:15; reference:arachnids,212;
reference:cve,1999-0532; classtype:attempted-recon;
sid:255; rev:11;)
```

Content generalised Rule 255:

```
Signature alert tcp $EXTERNAL_NET any -> $HOME_NET 53
(msg:"DNS zone transfer TCP"; flow:to_server,established;
content:"|00 00 |?||"; offset:15; reference:arachnids,212;
reference:cve,1999-0532; classtype:attempted-recon;
sid:255; rev:11;)
```

The DNS zone transfer attack exploits a buffer overflow in BIND version 4.9 releases prior to BIND 4.9.7 and BIND 8 releases prior to 8.1.2. An improperly or maliciously formatted inverse query on a TCP stream destined for the named service can crash the named server or allow an attacker to gain root privileges.

Generalisations of rules 323, 530, 1201 and 1378: These four generalised rules correctly pick up attacks. These attacks had already been spotted with the original SNORT rules. However, the additional packets found with the generalised rules provide the system administrator with better insight into the attacks by highlighting additional events that might be of interest.

We will use rule 1201 (ntinfoscan) as an example of how this occurred. The original rule reads:

```
alert tcp $HTTP_SERVERS $HTTP_PORTS -> $EXTERNAL_NET any
(msg:"ATTACK-RESPONSES 403 Forbidden";
flow:from_server,established; content:"HTTP/1.1 403";
depth:12; classtype:attempted-recon; sid:1201; rev:7;)
```

This was turned into the following generalised rule:

```
alert tcp $HTTP_SERVERS $HTTP_PORTS -> $EXTERNAL_NET any
(msg:"ATTACK-RESPONSES 403 Forbidden";
flow:from_server,established; content:"HTTP/1.|?| 403";
depth:12; classtype:attempted-recon; sid:1201; rev:7;)
```

Using the original rule alone, Snort did not pick up the additional packets/events because it was looking for HTTP/1.1, whereas the attacker used HTTP/1.0. Here is a part of some of the offending events:

```
GET /scripts/ HTTP/1.0
HTTP/1.0 403 Access Forbidden (Read Access Denied - This
Virtual Directory does not allow objects to be read.)
Content-Type: text/html
<body><h1>HTTP/1.0 403 Access Forbidden (Read Access Denied
- This Virtual Directory does not allow objects to be
read.)
</h1></body>
```

Finally Rule 1377: The original rule 1377 looks for two pieces of content '~' and '['. The generalised rule looks for only '[' and finds this because '[' is used to describe the beta version number as someone uses a beta version of wu ftp. This could be of interest to the system administrator as the use of beta-software might be against policy as is potentially introduces additional security risks.

Original Rule 1377:

```
alert tcp $EXTERNAL_NET any -> $HOME_NET 21 (msg:"FTP wu-
ftp bad file completion attempt [";flow:to_server,
established; content:"~"; content:"["; distance:1;
reference:bugtraq,3581; reference:bugtraq,3707;
reference:cve,2001-0550; reference:cve,2001-0886;
classtype:misc-attack; sid:1377; rev:14;)
```

Generalised Rule 1377:

```
alert tcp $EXTERNAL_NET any -> $HOME_NET 21 (msg:"FTP wu-
ftp bad file completion attempt [";flow:to_server,
established; content:"|?|"; content:"["; distance:1;
reference:bugtraq,3581; reference:bugtraq,3707;
reference:cve,2001-0550; reference:cve,2001-0886;
classtype:misc-attack; sid:1377; rev:14;)
```

Part of the actual event:

```
220 hobbes FTP server (Version wu-2.4.2-academ[BETA-15](1)
Sat Nov 1 03:08:32 EST 1997) ready. USER anonymous 331
Guest login ok, send your complete e-mail address as
password.
```

## 6.2 Content Generalisation or Catching Variants of the BugBear Virus

To see the potential benefit of content rule generalisation, let us consider a specific real-life example using the following generalised content rules against the BugBear Trojan virus. First, the SNORT rule shown below was created to identify a set of byte code within the virus. However, the BugBear virus (correctly known as the W32.Bugbear.B@mm worm) creates variations of itself as it spreads. We had both a '.scr' and a '.pif' variation to test against, but only the .scr variant was identified by the original rule. The .pif variation is as dangerous as the original and spreads just as quickly.

```
alert tcp any any -> any any (msg:Possible BugBear B
Attack; content:|3b 63 e7|; dsize:>21;)
```

Applying the generalised content FuzzRule program to the SNORT rule created three variations. Using these generalised variations, a match was then made against the .pif virus variation that did not previously escape detection. The generalised rules are shown below:

```
alert tcp any any -> any any (msg:Possible BugBear B Attack
FuzzRuleId cor(\'||?| 63 e7|\'); content:||?| 63 e7|;
regex; dsize:>21;)

alert tcp any any -> any any (msg:Possible BugBear B Attack
FuzzRuleId cor(\'|3b |?| e7|\'); content:|3b |?| e7|;
regex; dsize:>21;)

alert tcp any any -> any any (msg:Possible BugBear B Attack
FuzzRuleId cor(\'|3b 63 |?||\'); content:|3b 63 |?||;
regex; dsize:>21;)
```

## 7 Summary and Conclusions

In this paper we showed how, using simple generalisation, alert rules can be modified to show up new variants of old attacks. Using this method, we were able to identify rules that had too stringent criteria and also found new variants of a known Trojan.

Currently, only the surface has been scratched regarding generalised NIDS rule matching and it is difficult to make any definitive conclusions. However, some of the more unusual matches against generalised rules have shed light on how generalisation may aid SNORT, or indeed any NIDS, in finding undefined attacks. The techniques researched, developed and analysed have brought up a large number of false alerts. Once these alerts are eliminated, some potentially interesting alerts shine through.

Further investigation is required to determine fully how effective generalisation can be. For instance, it is important to work out how to distinguish more automatically false positives alerts from genuine new alerts generated by generalised rules. From the results and analysis in this paper, it seems that in particular applying generalisation to the content and uricontent SNORT rule parameters should be investigated further.

One hypothesis as to why applying generalisation to the uricontent option string appears more useful is that URI (e.g. web page address) strings could easily vary across attacks. An attack involving a URI string may have the same effect if a slightly different directory name is used, especially where standard directory names may vary across web server installations.